\documentclass[nohyperref]{article}

\usepackage{microtype}
\usepackage{graphicx}
\usepackage{subfigure}
\usepackage{booktabs} 

\usepackage{hyperref}

\usepackage[accepted]{icml2022}

\usepackage{amsmath}
\usepackage{amssymb}
\usepackage{mathtools}
\usepackage{amsthm}

\usepackage[capitalize,noabbrev]{cleveref}

\theoremstyle{plain}

\theoremstyle{definition}

\theoremstyle{remark}

\usepackage{color, colortbl}
\definecolor{Gray}{gray}{0.9}

\usepackage{amssymb}
\usepackage{pifont}
\newcommand{\cmark}{\ding{51}}%
\newcommand{\xmark}{\ding{55}}%
\usepackage{tablefootnote}
\usepackage{ulem}

\usepackage{algpseudocode}
\usepackage[titlenumbered, ruled,vlined]{algorithm2e}
\usepackage{amsmath}
\usepackage{amsthm}
\usepackage{bbm}
\usepackage{mathtools}

\DeclareMathAlphabet{\pazocal}{OMS}{zplm}{m}{n}
\newcommand{\unif}{\pazocal{U}}

\usepackage[textsize=tiny]{todonotes}

\newcommand{\std}[1]{\tiny{$\pm$ #1}}
\setuptodonotes{inline}    
\usepackage{adjustbox}
\usepackage{amsmath}

\newcommand*\ouralgo{{\sc SiFeR}}
\newcommand*\ouralgostr{\textbf{Si}eving \textbf{Fe}atures for \textbf{R}obust learning}

\icmltitlerunning{Overcoming Simplicity Bias in Deep Networks using a Feature Sieve}

\begin{document}

\twocolumn[
\icmltitle{Overcoming Simplicity Bias in Deep Networks using a Feature Sieve}




\begin{icmlauthorlist}
\icmlauthor{Rishabh Tiwari}{comp}
\icmlauthor{Pradeep Shenoy}{comp}
\end{icmlauthorlist}

\icmlaffiliation{comp}{Google Research India}

\icmlcorrespondingauthor{Pradeep Shenoy}{shenoypradeep@google.com}

\icmlkeywords{Machine Learning, ICML}

\vskip 0.3in
]



\printAffiliationsAndNotice{}  

\begin{abstract}

Simplicity bias is the concerning tendency of deep networks to over-depend on simple, weakly predictive features, to the exclusion of stronger, more complex features. This is exacerbated in real-world applications by limited training data and spurious feature-label correlations, leading to biased, incorrect predictions. We propose a direct, interventional method for addressing simplicity bias in DNNs, which we call the \textit{feature sieve}. We aim to automatically identify and suppress easily-computable spurious features in lower layers of the network, thereby allowing the higher network levels to extract and utilize richer, more meaningful representations. We provide concrete evidence of this differential suppression \& enhancement of \textit{relevant} features on both controlled datasets and real-world images, and report substantial gains on many real-world debiasing benchmarks (11.4\% relative gain on Imagenet-A; 3.2\% on BAR, etc). Crucially, we do not depend on prior knowledge of spurious attributes or features, and in fact outperform many baselines that explicitly incorporate such information. We believe that our \textit{feature sieve} work opens up exciting new research directions in automated adversarial feature extraction and representation learning for deep networks.

\end{abstract}

\section{Introduction}

\begin{figure}[!t]
\vskip 0.2in
\begin{center}
\centerline{
    \includegraphics[width=.65\linewidth]{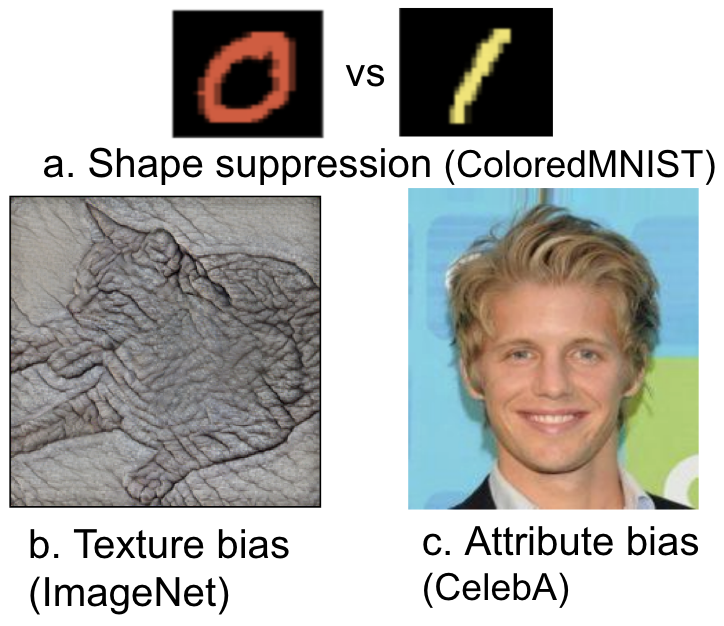}
}
\caption{Simplicity bias and spurious features. a) DNNs focus on color to the exclusion of shape when both are predictive. b) Image misclassified as elephant due to overdependence on texture features (adapted from~\cite{geirhos2018imagenet}). c) Classifiers mislabel blond-haired male faces as female.
}
\label{fig:datasets_examples}
\end{center}
\vskip -0.2in
\end{figure}

Deep networks are known to be vulnerable to a number of failure modes; in particular, \textit{simplicity bias} is the tendency of DNNs to prioritize weak predictive features over stronger, more difficult-to-extract features~\cite{shah2020pitfalls}. This bias has been studied analytically~\cite{pezeshki2021gradient} as well as empirically using natural images (texture bias~\cite{geirhos2018imagenet}) and carefully controlled synthetic datasets~\cite{hermann2020shapes} that independently manipulate feature complexity and predictive power. Such learning biases have significant real-world consequences too, resulting for instance in biased decision-making in AI-assisted workflows for face recognition, healthcare, credit rating, etc. \cref{fig:datasets_examples} illustrates the idea behind simplicity bias, and some real-world consequences. As a result, much recent work aims to \textit{debias} neural network models via a variety of approaches to achieve more equitable outcomes~\cite{mehrabi2021survey, zafar2017fairness, dwork2012fairness, russell2017worlds}.

Previous approaches towards debiasing DNNs include data manipulation via augmentation \& adversarial training~\cite{duboudin2022look,niu2022roadblocks}, data reweighting~\cite{nam2020learning}, multiple training environments~\cite{arjovsky2019invariant, pmlr-v162-zhou22e}, robust learning~\cite{pezeshki2021gradient}, and fairness objectives~\cite{li2022discover}. Other researchers have proposed diversity-enhanced ensembles~\cite{kim2022learning, teney2022evading,niu2022roadblocks} and architecture optimization \cite{bai2021ood}.

\begin{figure*}[!th]
\vskip 0.2in
\begin{center}
\centerline{
    \includegraphics[height=1.4in]{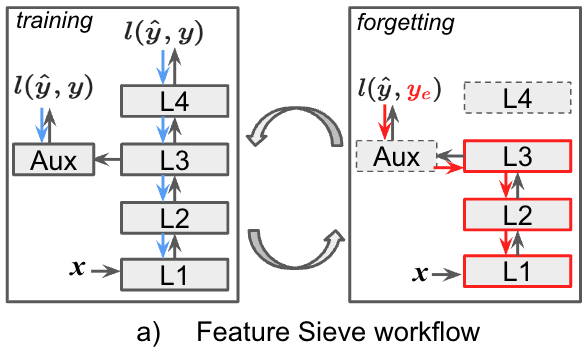}
    \hspace*{0.1in}
    \includegraphics[height=1.5in]{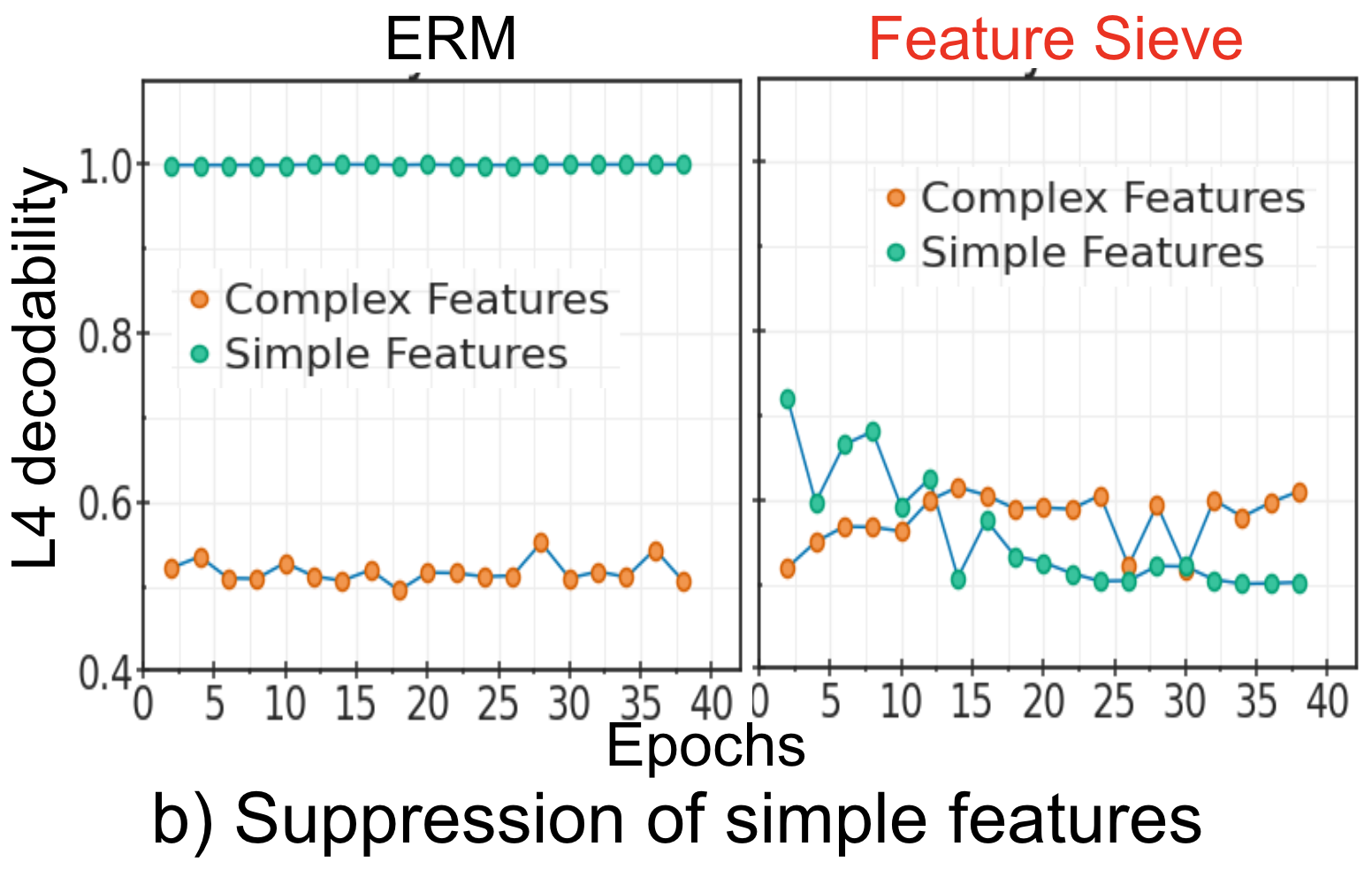}
     \hspace*{0.1in}
    \includegraphics[height=1.5in]{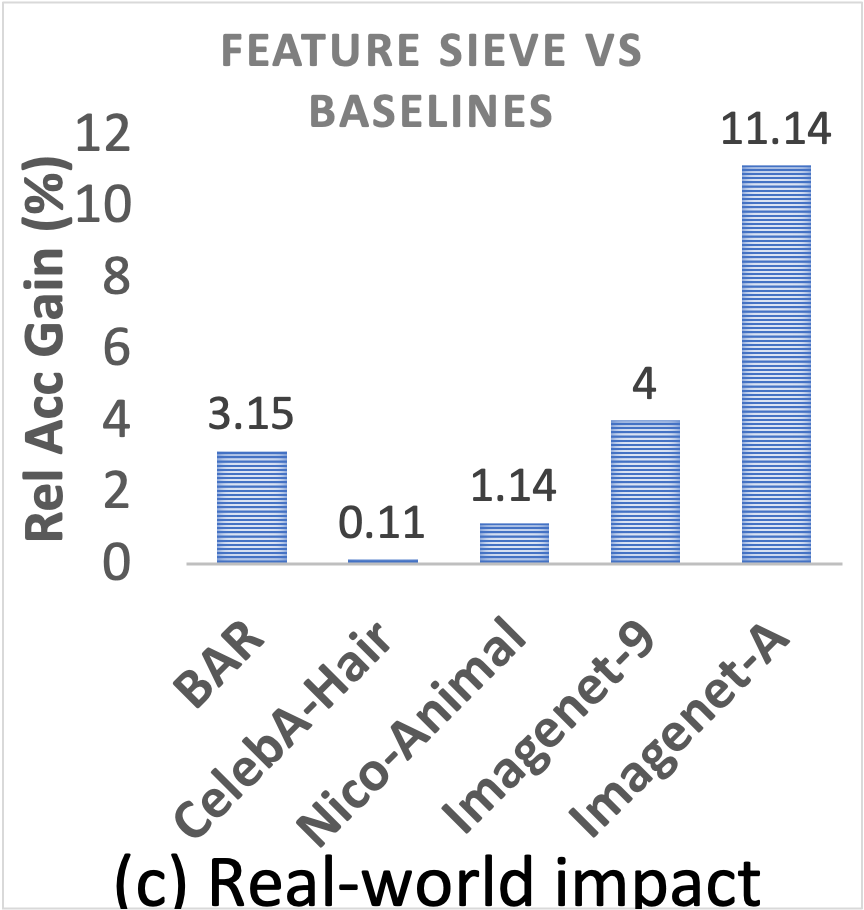}
}
\caption{\ouralgo\ workflow and results. a) We use an auxiliary network to alternately identify predictive features and erase them \textit{only at lower network layers}. By positioning the auxiliary network at different depths, we control the complexity of erased features. See \cref{sec:methods} for details. b) Our approach successfully suppresses digit and \textit{enhances CIFAR decodability} at higher layers for CIFAR\_MNIST dataset. c) We show significant gains over other approaches on many real-world debiasing benchmarks. 
}
\label{fig:workflow_gains}
\end{center}
\vskip -0.2in
\end{figure*}

We propose a novel, direct approach towards addressing simplicity bias in neural networks: an adversarial \textit{learning challenge} that forces the network to learn sophisticated feature representations. We refer to this learning challenge as a \textit{feature sieve}, and enforce it through the use of an auxiliary network (\cref{fig:workflow_gains}a;b). Our primary intuition is that simple features are computable early in the neural network, and proliferate throughout the deeper layers, thereby hindering the learning of complex features. We therefore propose to use the auxiliary network to alternately \textit{predict} labels using available features at some intermediate level (i.e., identify simple predictive features), and \textit{erase} those features from the early layers of the network, using a ``forgetting loss'' (see \cref{sec:methods} for details). Critically, our proposal does not depend on any specific definition or complexity class of ``simple features'', and instead automatically adapts to data characteristics using generalization error estimates.

We explicate our approach and its inner workings using experiments on controlled datasets (CMNIST, CIFAR\_MINST), and demonstrate its practical value on real-world debiasing benchmarks including BAR~\cite{nam2020learning}, CelebA~\cite{liu2018large}, NICO~\cite{he2021towards}, ImageNet-9~\cite{xiao2020noise} and ImageNet-A~\cite{hendrycks2021nae}; in nearly all experiments we show substantial gains over other competitive approaches. \cref{fig:workflow_gains}c provides a quick visual summary of our findings.

Summing up, we propose \ouralgo: \ouralgostr, a novel approach towards mitigating simplicity bias, thereby debiasing neural networks from spurious correlations in data. Our contributions are listed below.

\begin{itemize}
    \item We propose and formalize the idea of a feature sieve for mitigating simplicity bias, and provide an automated learning recipe to control feature complexity based on validation set.
    \item We show, using controlled datasets, the effectiveness of our approach in enhancing the decodability of complex features. We also demonstrate the customizability of our approach--our work is not restricted only to suppressing ``simple'' features, but is more broadly a controllable feature tradeoff tool.
    \item We show significant gains in debiasing classifiers on real-world datasets: 3.2\%, 4\%, 11.1\% relative gains over baselines on BAR, ImageNet-9, ImageNet-A (\cref{fig:workflow_gains}c). Crucially, we \textbf{do not use foreknowledge of biased features / input dimensions} in obtaining these results, unlike many of the baselines we outperform.
    \item Finally, we show using feature importance visualizations that \ouralgo\ is able to correctly identify important visual features of a scene, while suppressing irrelevant but spuriously-label-correlated background features (\cref{fig:gradcam}); this underscores the relevance of \ouralgo\ to real-world feature understanding.
\end{itemize}

We hope that our work with \ouralgo\footnote{Code available at https://github.com/google-research/google-research/sifer} encourages further work in designing interesting computational barriers for neural networks; by automating the extraction and combination of diverse features ordered by complexity and predictive power, we could make significant progress towards the debiasing of machine learning models.

\section{Related Work}

\subsection{Simplicity Bias}
\citet{shah2020pitfalls} showed that neural networks trained with SGD are biased to learn the simplest predictive features in the data while ignoring others. Numerous studies have attempted to investigate the correlation and impact of such shortcuts, yielding a wealth of intriguing findings \cite{nagarajan2020understanding, hermann2020shapes}. 

\subsection{Debiasing Spurious Correlations}
Unlike our work, the majority of previous work on mitigating simplicity bias uses explicit biased-attribute labels \cite{kim2019learning, li2019repair, sagawa2019distributionally, teney2020unshuffling, krueger2021out, bai2020decaug} in their debiasing recipes. This reduces their practicality since both identifying, and manually labeling biased instances and dimensions in real-life data are significant barriers. Only recently, the focus has shifted towards debiasing without using explicit attribute labels ~\cite{teney2022evading, kim2022learning, niu2022roadblocks, shrestha2022occamnets, nam2020learning}. Here we discuss different technical approaches used by previous work in both of the above directions:

\textit{Alternate Networks:} LfF~\cite{nam2020learning} and LWBC~\cite{kim2022learning} initially train a prejudiced network and try to debias the second network by focusing on samples that go against the bias. \\
\textit{Ensemble:} LWBC~\cite{kim2022learning} and ESB~\cite{teney2022evading} both create a classifier ensemble; the former enforces debiasing via reweighting of training instances, and the latter incorporates a diversity constraint in the ensemble. \\
\textit{Architecture Design:} NAS-OoD~\cite{bai2021ood} adds an OOD generalization criterion to network architecture search training to select inherently more robust network architectures. OccamNet~\cite{shrestha2022occamnets} adds a few inductive biases in the network--for instance, explaining the dataset with simple hypotheses and bounded network depth, and applying spatial localization assumptions about unbiased (visual) features in order to filter spurious features. \\
\textit{Multiple Environments:} IRM~\cite{arjovsky2019invariant} uses the theory of causal bayesian networks to find an invariant feature representation using multiple training environments with different bias correlations. REx~\cite{krueger2021out} tries to improve on the worst linear combinations of risks from different training environments. CaaM~\cite{wang2021causal} learns causal attention by partitioning the data on-the-go to break correlation with bias. \\
\textit{Augmentations:} DecAug~\cite{bai2020decaug} proposed a semantic augmentation and feature decomposition approach to disentangle context features from category related features. \citet{niu2022roadblocks} adds adversarial augmentations to the image while training to avoid over-reliance on spurious visual cues. This work is conceptually closest to our work, in that it builds an ensemble where previous components compete with a new classifier to encourage it to learn diverse hypotheses. Our approach directly addresses the competitive development of features within a network (the ``heart'' of the simplicity bias challenge); we also outperform them on the BAR dataset~\cite{nam2020learning} (\cref{bar-table}), while being more computationally parsimonious.

\section{\ouralgo: a Feature Sieve for Bias Mitigation} \label{sec:methods}
\subsection{Preliminaries \& Intuition}
We start from the assumption that simple features are (by definition) quickly learned, available early in the neural network stack (i.e., in layers closer to the input), and more easily proliferate throughout the subsequent layers (see e.g.,~\citet{hermann2020shapes} for substantial supportive evidence for these assumptions). Further, the ubiquitous presence of simple features actively prevents acquisition of more complex hypotheses by subsequent NN layers, due to the so-called simplicity bias inherent in NN training methods--see e.g.,~\citet{shah2020pitfalls, pezeshki2021gradient} for theoretical results supporting these claims.

Thus, our primary goal is to \textit{identify and actively suppress} simple / spurious predictive features, so as to create room for the learning of complex predictive features at higher layers of the NN--an approach we refer to as a ``feature sieve''.  

We include another key consideration in the design of our approach: do not leverage any \textit{a priori} information of simple features, or even the function class / degree of complexity of simple features. To support this design goal, we a) build into our design the knobs that control tradeoffs between simpler- and more complex-to-compute features, and b) focus on reducing generalization error as the objective in setting these knobs. This allows us to not only automatically discover useful tradeoffs, but also to ensure that our trained classifiers are overall more accurate than standard baselines.

As a final remark, we note that the distinctions between simple / complex, spurious / accurate, early-layer / late-layer, and early-acquisition / late-acquisition are likely substantially more nuanced than a simple one-to-one correspondence, even though they are often used interchangeably for ease of exposition. For instance, depending on the dataset, a ``simple'' feature may in fact be the best / most unbiased predictive feature. For this reason, too, depending upon generalization error for controlling the feature sieve is strongly preferred to the use of any stronger inductive bias along the dimensions mentioned above.

\subsection{The Alternating Identify-and-Erase Workflow}

\cref{fig:workflow_gains}(a) provides an overview of \ouralgo. Briefly,  we use an \textit{auxiliary network}, working  at an intermediate level of representation in the neural network, to identify predictive features (simple / spurious) in the representation, and subsequently to erase them at the lower layers of the primary network. This is a direct operationalization of our primary goal stated above. 

\textbf{Identifying simple features:}  The training of the primary network proceeds in conventional fashion via forward- and back-propagation (\cref{fig:workflow_gains}(a), left panel, black \& blue arrows respectively), with an additional auxiliary layer that learns to predict the label from an intermediate representation. Note that feedback from the auxiliary layer does not back-propogate to the main network. This is a conscious decision choice to force the auxiliary layer to learn from already-available features rather than create or reinforce them in the main network. By controlling the auxiliary network's capacity and the layer of the primary network to which it is attached, we can control the complexity of the predictive features it can identify.

\textbf{Applying the feature sieve:} We aim to \textit{erase} the identified features in the early layers of the neural network, by the combination of the following steps: a) The parameters of auxiliary layer ($\mathcal{A}$) are frozen, and only that portion of the main network ($\mathcal{M}_d$) which is before the auxiliary layer is kept trainable-- this is the region where we wish to ``forget'' the simple features, and b) We apply a \textit{forgetting loss} ($\mathcal{L}_f$) at the output layer of the auxiliary network. 
\begin{eqnarray}
\hat{\mathbf{y}}_{aux} &=& \mathcal{A}(\mathcal{M}_d(\mathbf{x}))\\
\mathbf{y}_{ep} &=& [\frac{1}{n},  \frac{1}{n}, ... ] \ \ \textit{(n\ entries)} \\
\mathcal{L}_f &=& \texttt{CE}(\hat{\mathbf{y}}_{aux}, \mathbf{y}_{ep})
\end{eqnarray}
where $\mathbf{x}$, $\mathbf{y}_{ep}$, $\hat{\mathbf{y}}_{aux}$ and $n$ represent input images, a pseudo-label with uniform probability  across classes, the prediction from auxiliary layer, and number of classes respectively.


\textbf{Iterative optimization:} A challenge is that this process of identification and sieving is dynamic in nature; in particular, the two steps may interfere with each other. In order to handle this challenge, we \textit{interleave} the two steps such that each forgetting step happens after regular intervals of some minibatch iterations ($\mathcal{F}$) which is treated as a hyperparameter selected using the validation set.

The entire learning recipe is summed up in Algorithm \ref{algo1}

\begin{algorithm}[h]
\SetAlgoLined
\SetKwInOut{Input}{Input}
\SetKwInOut{Hparams}{Hparams}
\SetKwInOut{Output}{Output}
\Input{Pretrained Model Weights $\mathbf{W}$;\\ 
training data $\mathcal{D}$; training iters $N$}
\Hparams{Aux Depth $\mathcal{A}_D$; Aux Position $\mathcal{A}_P$; \\ main\_lr\_weight $\alpha_1$; aux\_lr\_weight $\alpha_2$; \\
aux\_forget\_weight $\alpha_3$; forget\_after\_iters $\mathcal{F}$}
\Output{robust model weights $\mathbf{W}$}
\For {$k = 1 \hdots N$} 
{
    $(\mathbf{x}, \mathbf{y}) \gets \texttt{sample}(\mathcal{D})$ \\
    $\hat{\mathbf{y}}, \hat{\mathbf{y}}_{aux}  \gets \texttt{Forward\_with\_aux}(\mathbf{x}, \mathcal{A}_D, \mathcal{A}_P, \mathbf{W})$ \\
    $\mathcal{L}_1 \gets \Call{CE}{\hat{\mathbf{y}}, \mathbf{y}}$ \\
    $\mathcal{L}_2 \gets \Call{CE}{\hat{\mathbf{y}}_{aux}, \mathbf{y}}$ \\
    $\mathcal{L}_f \gets \Call{CE}{\hat{\mathbf{y}}_{aux}, \unif}$ \\
    $\mathcal{L} \gets \alpha_1 \mathcal{L}_1 + \alpha_2 \mathcal{L}_2$ \\
    \If{k \% $\mathcal{F}$ == 0}
        {$\mathcal{L} \gets \mathcal{L} + \alpha_3 \mathcal{L}_f$}
    $\nabla{\mathbf{W}} \gets \texttt{Backward}(\mathcal{L})$ \\ 
    $\mathbf{W} \gets \texttt{OptimizeStep}(\nabla\mathbf{W})$
}
\caption{\ouralgo: Mitigating simplicity bias}
\label{algo1}
\end{algorithm}

\subsection{Controllability of the Feature Sieve} \label{sec:control}

As remarked earlier, we aim to automatically discover notions of and tradeoffs between so-called simple and complex features, as relevant for the specific dataset at hand. The feature sieve approach described here allows for many mechanisms to control this discovery \& tradeoff. The primary parameters are the position \& depth of the auxiliary network ($\mathcal{A}_P, \mathcal{A}_D$) which implicitly control the function complexity of the features available for discovery by the auxiliary network; and the auxiliary forgetting weight $\alpha_3$, which controls the degree to which the discovered features are suppressed. The interleaving of the feature identifying \& feature sieving steps is controlled by the parameter $\mathcal{F}$--again, based on the specific dataset and the nature of the features contained, this controls the dynamics of the training procedure. 

Finally, we set these hyperparameters based on the goal of minimizing validation error--this ensures not only that the parameters are chosen using unbiased estimates of generalization, but also that at a minimum, we perform better than the standard training baseline (which, as the trivial solution of not-forgetting, is included in the search space for the feature sieve).

\section{Experiment Setup}\label{sec:setup}

\subsection{Datasets for Studying Simplicity Bias}
\textbf{CMNIST}: Colored-MNIST is a 2-class synthetic dataset used to study simplicity bias. We use digits $0$ \& $1$ respectively from the MNIST dataset, with an added color channel (red for $0$ images, green for $1$). 

\textbf{CIFAR-MNIST}: Similar to CMNIST, this binary classification dataset has paired-composite images--Class $0$ pairing MNIST  $0$s with CIFAR automobiles, and Class $1$ pairing MNIST $1$s with CIFAR truck images.

Both datasets contain perfectly predictive simple \textit{and} complex features; by training a classifier and manipulating the test set to break one of these correlations, one can examine which features are being used by the trained classifier.

\subsection{Real-World Debiasing Benchmarks}
\textbf{BAR:} Biased Activity Recognition \cite{nam2020learning} is a real-world image benchmark for classifying human actions (images) into 6 classes; each training image contains spurious correlations with background features (e.g., rocks with climbing). The test set contains the same set of actions but with different backgrounds (e.g., ice with climbing). The training data has no bias-conflicting examples (i.e., examples which violate the spurious correlation), which makes this a challenging benchmark.

\textbf{CelebA: } CelebA~\cite{liu2018large} contains human faces, each labelled with 40 attributes. Following \citet{kim2022learning}, we focus on predicting \texttt{HairColor}, an attribute heavily correlated with \texttt{Gender} in the dataset. Specifically, most CelebA images with \texttt{blond-hair} (more than 99\% of them in the training set) are women.

\textbf{NICO:} NICO \cite{he2021towards} is a real-world benchmark for out-of-distribution robustness. Following \citet{bai2020decaug}, we used its Animal subset containing 10 object classes and 10 context labels. The training set only contains 7 contexts for each object class while the validation and test set contains 3 extra unseen contexts (total 10). Unlike the majority of the baselines, we don't use context label attributes in train, validation, or test. 

\textbf{ImageNet-9:} ImageNet-9 \cite{xiao2020noise} is a subset of ImageNet \cite{imagenet_cvpr09} containing 9 super classes. It has been established that this subset has a spurious correlation between object labels and image texture. We followed the setting used by \cite{kim2022learning} and \cite{bahng2020learning} for creating train and val split. We report the average accuracy on the validation split.

\textbf{ImageNet-A: } ImageNet-A~\cite{hendrycks2021nae} contains handpicked real-world images misclassified by models trained on ImageNet. Since these misclassifications are due to over-reliance on spurious features like color\&texture, we use this dataset for evaluating models trained on ImageNet-9 as a robustness challenge (i.e., OOD test set).


\subsection{Training Procedure \& Metrics}\label{sec:decode}
For all our real-world experiments we consistently used ResNet-18, an auxiliary layer that uses the same layer structure as of BasicBlock of ResNet with varying depth, optimized using SGD optimizer with a fixed learning rate of 0.001. For real-world experiments the model is loaded with ImageNet pre-trained weights. We repeat the experiments with 5 different random seeds and report the mean and std deviation of results. Refer Appendix \ref{app-training_details} for more details.

\textbf{Choice of Validation Set:} For BAR, since there is no validation data provided, we study it under two settings. In the first, we use 20\% of images from the test set and call it OOD-validation. In the second setting, which is  harder and more realistic, we use 20\% images from the train set, calling it In-Domain (ID) validation. For NICO-Animal, CelebA Hair and ImageNet-9, we use the already supplied validation data.  Table \ref{dataset-composition-table} shows the percentage of ``bias-conflicting'' examples, i.e., examples that violate the spurious feature correlation or training domain, for each portion of each dataset. Note that BAR-ID val setting, NICO and ImageNet-9/ImageNet-A experiments do not have any bias conflicting examples in the train set, and methods that rely on attribute labels and/or reweighting of training data will perform poorly on them.

\begin{table}[t]
\caption{Composition of conflicting examples in different datasets.}
\label{dataset-composition-table}
\begin{center}
\begin{tabular}{lcccc}
\toprule
\textbf{Dataset}   & \multicolumn{3}{l}{\textbf{\%  Conflict Examples}}              & \textbf{ hparam goal} \\
          & \multicolumn{1}{l}{Train} & \multicolumn{1}{l}{Val} & Test &                               \\ \midrule
BAR-ID val      & 0                         & 0                       &  100    & Avg Acc                   \\
BAR-OOD val       & 0                         & 100                       &  100    & Avg Acc                   \\
CelebA    & 0.8                       & 0.9                     &  0.9    & Unbiased Acc                  \\
NICO      & 0                         & 10                      &  10    & Avg Acc                   \\
IN-9/IN-A & 0                         & 0                       &  100    & Avg Acc    \\ \bottomrule              
\end{tabular}
\end{center}
\end{table}

\textbf{Evaluation Metrics:} \texttt{Accuracy} means average accuracy on all examples. \texttt{Unbiased} means accuracy averaged over each label-context group. This metric is more fair when there is a huge imbalance between the groups. \texttt{Conflicting} means accuracy only on the bias conflicting examples.

We used \texttt{Accuracy} for BAR, NICO and Imagenet-9 / ImageNet-A and \texttt{Unbiased} for CelebA-Hair dataset as the performance metric on validation data for hyperparameter search and early stopping (\cref{dataset-composition-table}).

\textbf{Feature decodability:} To measure the ``decodability'' of a chosen feature in a classifier at a given layer, we freeze the classifier and train a linear decoder on its network representation at the specified layer. The decoder is trained on validation data, with each instance being assigned as label the value of its feature. For example, to check decodability of shape in a CMNIST classifier, input instances are assigned the label of the shape they contain, while ignoring their color. This decoder's accuracy is then reported on the test set.

\section{Results}
\begin{figure*}[!th]
\vskip 0.2in
\begin{center}
\includegraphics[width=\linewidth]{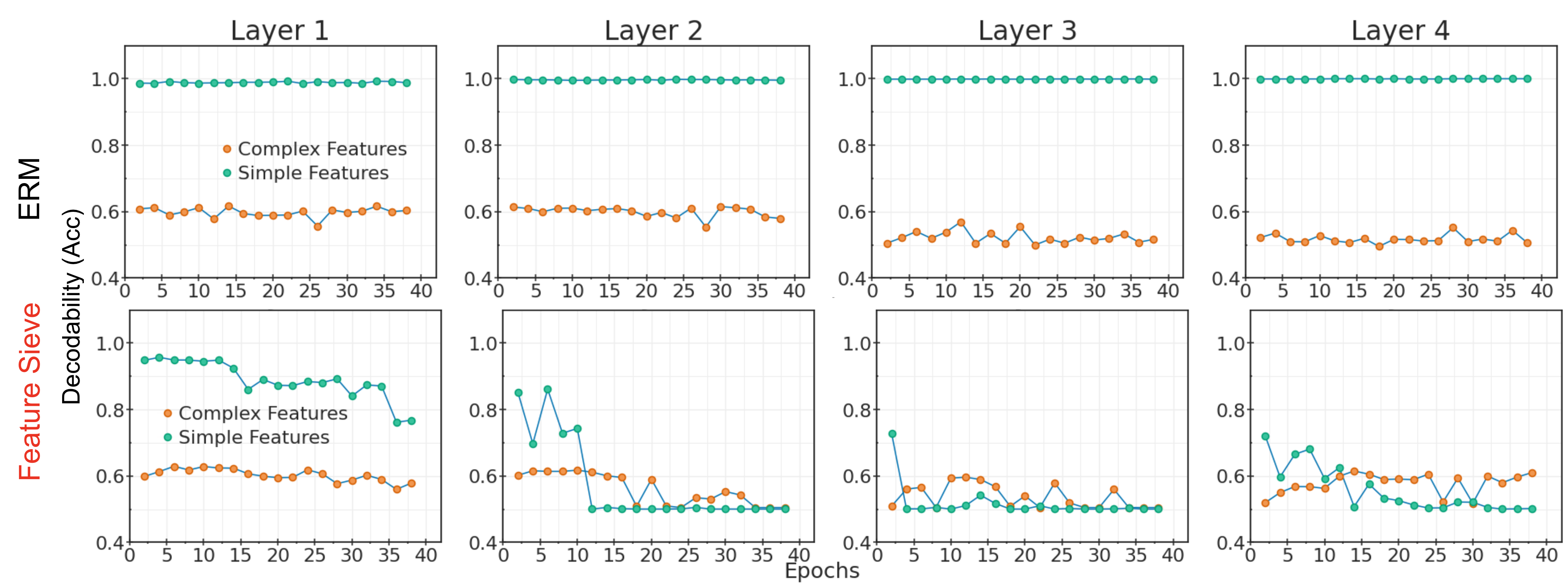}
\vskip -0.2in
\caption{Decodability of Simple (MNIST) and Complex (CIFAR) features across layers of ResNet-50 with a) Normal ERM training b) with \ouralgo}
\label{fig:cifar_mnist_decodability}
\end{center}
\vskip -0.2in
\end{figure*}

\begin{table*}[t]
\caption{Feature Controllability. 
}
\label{feature-controllability-table}
\begin{center}
\begin{tabular}{cc|cc|cc}
\toprule
DataSet              & Target Feature  & \multicolumn{2}{c|}{\ouralgo\ (Ours)}                                                & \multicolumn{2}{c}{ERM}                                                 \\
                     &                 & SR                                 & CR                                  & SR                                 & CR                                 \\ \midrule
\multicolumn{1}{c}{CMNIST}      & Complex (Digit)      & \cellcolor[HTML]{59BC8C}99.54$\pm$0.19 & \cellcolor[HTML]{DDF1E7}58.14$\pm$10.69 & \cellcolor[HTML]{E0F3EA}56.96$\pm$6.59 & \cellcolor[HTML]{70C69C}92.21$\pm$3.92 \\
                                & Simple (Color)       & \cellcolor[HTML]{EFF9F4}52.44$\pm$1.22 & \cellcolor[HTML]{59BC8B}99.64$\pm$1.30  & \cellcolor[HTML]{F9FDFB}49.20$\pm$2.60 & \cellcolor[HTML]{63C093}96.27$\pm$0.99 \\ \midrule
\multicolumn{1}{c}{CIFAR\_MNIST} & Complex (CIFAR)      & \cellcolor[HTML]{CFECDE}62.37$\pm$4.62 & \cellcolor[HTML]{FAFDFC}48.93$\pm$1.92  & \cellcolor[HTML]{DDF1E7}58.14$\pm$1.60 & \cellcolor[HTML]{57BB8A}100        \\
                                & Simple (Digit)       & \cellcolor[HTML]{FFFFFF}47.17$\pm$0.14 & \cellcolor[HTML]{58BC8B}99.83$\pm$0.29  & \cellcolor[HTML]{F9FDFB}49.20$\pm$2.60 & \cellcolor[HTML]{57BB8A}100     \\ \bottomrule  
\end{tabular}
\end{center}
\end{table*}

\subsection{Suppressing Simple Features}

We first studied the effectiveness of \ouralgo\ on targeted suppression of specific features. To do this, we experimented with the CIFAR\_MNIST dataset, which consists of composite pairings of MNIST \& CIFAR images, each fully predictive of the assigned label (see \cref{sec:setup} for more details). DNNs are known to entirely ignore the CIFAR feature on this training dataset--when the CIFAR component is randomized at test-time, accuracy is unaffected, but when the MNIST component is randomized, accuracy drops to chance. We refer to the MNIST component as the simple feature, and CIFAR as the complex feature.

\cref{fig:cifar_mnist_decodability} shows layerwise decodability (\cref{sec:decode}) of simple and complex features, tracked across epochs in the training process. We contrast standard training (top row) against \ouralgo\ (bottom row). 
Standard training overemphasizes the simple feature at higher layers, and ignores the complex feature. The complex CIFAR feature is in fact decodable to some extent in earlier layers of the ERM classifier, but is suppressed in later layers, due to the preponderance of the simple feature. In contrast, the auxiliary forgetting loss in \ouralgo\ effectively \textit{suppresses the simple feature} in the earlier layers, and thereby enhances the decodability of the complex feature in the higher layers. This shows that removing the availability of spurious simple features is a direct method of overcoming simplicity bias. 

These findings are more remarkable given that no prior knowledge of ``simple''/``complex'' features was used in any way whatsoever during training. \ouralgo\ organically discovers and suppresses the by-design simple feature purely through the use of the strategically placed auxiliary network, and its configuration via the training recipe.

\subsection{Feature Controllability using \ouralgo}\label{sec:controllability}
We described in \cref{sec:control} the various degrees of freedom in \ouralgo\ for identifying and suppressing features, and choosing them by the use of a validation set (i.e., using generalization error). This gives us a simple, powerful method for certain kinds of domain generalization, by simply using domain-shift data as the validation set. We demonstrate this capability by conducting studies on controlled datasets CMNIST \& CIFAR\_MNIST. In each, the training data were designed to have pairings of simple and complex features where both simple and complex features were fully predictive. We then trained a \ouralgo\ classifier for \textit{different choices of validation set}, representing which feature we actually wanted the classifier to focus on. This was achieved by randomizing the ``non-relevant'' feature in the validation dataset, and choosing all our hyperparameters based on that validation dataset. This represents a real-world scenario where small amounts of vetted data are available for optimization of a model, but (re)-labeling or manipulating all training data is infeasible. \cref{feature-controllability-table} shows the results comparing \ouralgo\ against ERM. \ouralgo\ shows higher accuracy for chosen features (higher diagonal terms) than for spurious features (off-diagonal terms), driven by the choice of the validation data. In contrast, ERM primarily focuses on the simple features, irrespective of the choice of validation set (higher second column numbers). Thus, our method is in fact able to focus on the relevant feature--be it simple or complex--in an easily controllable manner.

\begin{figure*}[!th]
\vskip 0.2in
\begin{center}
\includegraphics[width=\linewidth]{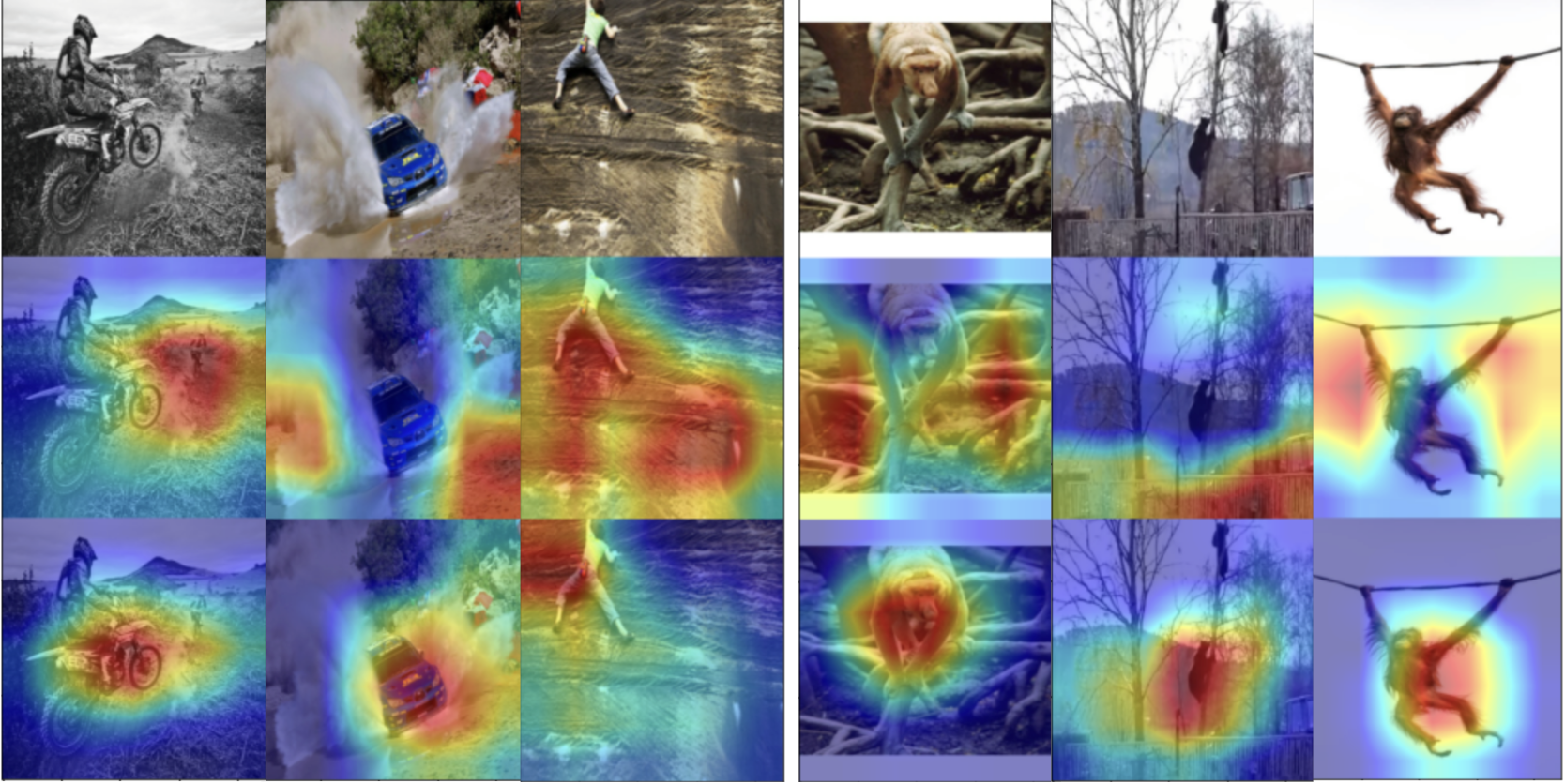}
\vskip -0.1in
\caption{Examples of \ouralgo's focus on \textit{relevant features} while suppressing irrelevant background information. Top row: Input images; Middle row: GRAD-CAM-derived feature importance visualizations for the ERM classifier; Bottom row: feature importance for \ouralgo.  First 3 columns from BAR~\cite{nam2020learning}, and last 3 columns from NICO-Animal \cite{he2021towards}. 
}
\label{fig:gradcam}
\end{center}
\vskip -0.2in
\end{figure*}

\subsection{Debiasing Real-World Datasets}


\begin{table}[t]
\caption{Classification Accuracy (\%) on test set of BAR Dataset.}
\label{bar-table}
\begin{center}
\begin{tabular}{lcc}
\toprule
\multicolumn{1}{c}{Method} & Used OOD Val             & \texttt{Accuracy}             \\ \midrule
                           & \multicolumn{1}{l}{}     & \multicolumn{1}{l}{} \\
ERM                        & {\color[HTML]{FF0000} \cmark} & 51.85\std{5.92}           \\
BiaSwap\tiny~\cite{kim2021biaswap}                    & {\color[HTML]{FF0000} \cmark} & 52.44                \\
LfF\tiny~\cite{nam2020learning}                        & {\color[HTML]{FF0000} \cmark} & 62.98\std{2.76}           \\
PGI\tiny~\cite{ahmed2021systematic}                        & {\color[HTML]{FF0000} \cmark} & 65.19\std{1.32}           \\
EIIL\tiny~\cite{creager2021environment}                       & {\color[HTML]{FF0000} \cmark} & 65.44\std{1.17}           \\
ESB\tablefootnote{ESB uses R50 architecture unlike other baseline which uses R18.}\tiny~\cite{teney2022evading}                        & {\color[HTML]{FF0000} \cmark} & 67.10\std{0.30}           \\
Roadblock\tiny~\cite{niu2022roadblocks}                        & {\color[HTML]{FF0000} \cmark} & 69.51\std{2.43}           \\
Debian\tiny~\cite{li2022discover}                     & {\color[HTML]{FF0000} \cmark} & 69.88\std{2.92}           \\
\rowcolor{Gray}
\ouralgo\ (Ours)                       & {\color[HTML]{FF0000} \cmark} & \textbf{72.08\std{0.38}}  \\ \midrule
ERM                        & {\color[HTML]{1D8E3E} \xmark} & 35.32\std{0.46}           \\
ReBias\tiny~\cite{bahng2020learning}                     & {\color[HTML]{1D8E3E} \xmark} & 37.02\std{0.26}           \\
LfF\tiny~\cite{nam2020learning}                        & {\color[HTML]{1D8E3E} \xmark} & 48.15\std{0.93}  \\ 
SSL+ERM\tiny~\cite{kim2022learning}                    & {\color[HTML]{1D8E3E} \xmark} & 60.88\std{0.80}           \\
LWBC\tiny~\cite{kim2022learning}                       & {\color[HTML]{1D8E3E} \xmark} & 62.03\std{0.74}           \\
ESB\tiny~\cite{teney2022evading}                       & {\color[HTML]{1D8E3E} \xmark} & 64.40\std{0.20}           \\
\rowcolor{Gray} 
\ouralgo\ (Ours)                       & {\color[HTML]{1D8E3E} \xmark} & \textbf{65.75\std{1.84}} \\
\bottomrule
\end{tabular}
\end{center}
\end{table}
\begin{table}[t]
\caption{Unbiased and Conflicting Accuracy metrics (\%) on Test set of CelebA Hair Dataset }
\label{celeba-table}
\begin{center}
\begin{adjustbox}{width=0.5\textwidth}
\begin{tabular}{lccc}
\toprule
\multicolumn{1}{c}{Method} & \begin{tabular}[c]{@{}c@{}}Spurious  \\ Attribs\end{tabular} & \texttt{Unbiased}        & \texttt{Conflict}     \\ \midrule
DRO\tiny{~\cite{sagawa2019distributionally}}                  & {\color[HTML]{FF0000} \cmark} & 85.43\std0.53           & 83.40\std0.67           \\
EnD\tiny~\cite{tartaglione2021end}                        & {\color[HTML]{FF0000} \cmark} & 91.21\std0.22           & 87.45\std1.06           \\
CSAD\tiny~\cite{zhu2021learning}                       & {\color[HTML]{FF0000} \cmark} & 89.36                & 87.53                \\ \midrule
ERM                        & {\color[HTML]{1D8E3E} \xmark} & 70.25\std0.35           & 52.52\std0.19           \\
LfF\tiny~\cite{nam2020learning}                        & {\color[HTML]{1D8E3E} \xmark} & 84.24\std0.37           & 81.24\std1.38           \\
SSL+ERM\tiny~\cite{kim2022learning}                    & {\color[HTML]{1D8E3E} \xmark} & 80.48\std0.91           & 66.79\std2.20           \\
LWBC\tiny~\cite{kim2022learning}                       & {\color[HTML]{1D8E3E} \xmark} & 88.90\std1.55           & 87.22\std1.14           \\
\rowcolor{Gray}
\ouralgo\ (Ours)                      & {\color[HTML]{1D8E3E} \xmark} & \textbf{89.00\std0.92}  & \textbf{88.04\std1.25} \\
\bottomrule
\end{tabular}
\end{adjustbox}
\end{center}
\end{table}

\begin{table}[t]
\caption{Classification Accuracy (\%) on test set of NICO Dataset. Most of the baselines (DecAug, DRO, etc) use spurious attribute labels for training, while we do not.}
\label{nico-table}
\begin{center}
\begin{tabular}{lcc}
\toprule
\multicolumn{1}{c}{Method} & \texttt{Accuracy}             \\ \midrule
                          &                                                                                         \\
ERM                                                      & 75.87                \\
IRM\tiny~\cite{arjovsky2019invariant}                                                                   & 59.17                \\
REx\tiny~\cite{krueger2021out}                                                                   & 74.31                \\
JiGen\tiny~\cite{carlucci2019domain}                                                                                    & 84.95                \\
Mixup\tiny~\cite{zhang2017mixup}                                                                                     & 80.27                \\
Cumix\tiny~\cite{mancini2020towards}                                                                                    & 76.78                \\
MTL\tiny~\cite{blanchard2021domain}                                                                                      & 78.89       \\
DANN\tiny~\cite{ganin2016domain}                                                                                    & 75.59                \\
CORAL\tiny~\cite{sun2016deep}                                                                                     & 80.27                \\
MMD\tiny~\cite{li2018domain}                                                                                       & 70.91       \\
DRO\tiny~\cite{sagawa2019distributionally}                                                                  & 77.61                \\
CNBB\tiny~\cite{he2021towards}                                                                                     & 78.16                \\
DecAug\tiny~\cite{bai2020decaug}                                                               & 85.23       \\
\rowcolor{Gray}
\ouralgo\ (Ours)                                                                & \textbf{86.20\std0.85}  \\ \midrule
NAS-OoD\tablefootnote{Architecture design optimization based method, hence unfair to compare directly against other methods.}\tiny~\cite{bai2021ood}                                                               & 88.72     \\
\bottomrule
\end{tabular}
\end{center}
\end{table}

\begin{table}[t]
\caption{Classification Accuracy (\%) on Validation set of ImageNet-9 and test set of ImageNet-A.}
\label{imagenet9-table}
\begin{center}
\begin{adjustbox}{width=0.5\textwidth}
\begin{tabular}{lccc}
\toprule
\multicolumn{1}{c}{Method} & \begin{tabular}[c]{@{}c@{}}Spurious \\ Attribs\end{tabular} & ImageNet-9          & ImageNet-A          \\ \cline{3-4} 
                           & \multicolumn{1}{l}{{\color[HTML]{FF0000} }}                         & \texttt{Accuracy}      & \texttt{Accuracy}            \\ \midrule
StylisedIN\tiny~\cite{geirhos2018imagenet}                 & {\color[HTML]{FF0000} \cmark}                                            & 88.4\std0.5            & 24.6\std1.4            \\
LearnedMixin\tiny~\cite{clark2019don}               & {\color[HTML]{FF0000} \cmark}                                            & 64.1\std4.0            & 15.0\std1.6            \\
RUBi\tiny~\cite{cadene2019rubi}                       & {\color[HTML]{FF0000} \cmark}                                            & 90.5\std0.3            & 27.7\std2.1            \\ \midrule
ERM                        & {\color[HTML]{1D8E3E} \xmark}                                            & 90.8\std0.6            & 24.9\std1.1            \\
BagNet18\tiny~\cite{brendel2019approximating}                   & {\color[HTML]{1D8E3E} \xmark}                                            & 67.7\std0.3            & 18.8\std1.15           \\
ReBias\tiny~\cite{bahng2020learning}                      & {\color[HTML]{1D8E3E} \xmark}                                            & 91.9\std1.7            & 29.6\std1.6            \\
LfF\tiny~\cite{nam2020learning}                        & {\color[HTML]{1D8E3E} \xmark}                                            & 86.00               & 24.60       \\
CaaM\tiny~\cite{wang2021causal}                       & {\color[HTML]{1D8E3E} \xmark}                                            & 95.70               & 32.80                \\
SSL+ERM\tiny~\cite{kim2022learning}                    & {\color[HTML]{1D8E3E} \xmark}                                            & 94.18\std0.07          & 34.21\std0.49          \\
LWBC\tiny~\cite{kim2022learning}                       & {\color[HTML]{1D8E3E} \xmark}                                            & 94.03\std0.23          & 35.97\std0.49          \\
\rowcolor{Gray}
\ouralgo\                   & {\color[HTML]{1D8E3E} \xmark}                                            & \textbf{97.78\std0.12} & \textbf{39.98\std0.81} \\
\bottomrule
\end{tabular}
\end{adjustbox}
\end{center}
\end{table}

Our method outperforms baselines on four different real world datasets--BAR~\cite{nam2020learning}, CelebA Hair~\cite{liu2018large}, NICO~\cite{he2021towards} and Imagenet-A~\cite{hendrycks2021nae}, by large margins (upto 11\%, see \cref{fig:workflow_gains}c for a quick summary). Critically, we chose in all our experiments to \textit{not use any knowledge} of which attribute labels are considered spurious in each dataset--this is because in real-world scenarios, it is difficult to know in advance which attributes may end up containing biased information, or to label data according to those attributes in order to do targeted debiasing of models. Nevertheless, we outperform the other baselines, including many that do use attribute labels as part of their training procedure.

\textbf{Mitigating Spurious Correlations:} Biased Activity Recognition (BAR) and CelebA Hair Dataset represent background and gender bias in real life. In the BAR training set, human activity (image categories) is spuriously correlated with the background in which those activities are performed; in CelebA Hair, hair color is strongly correlated with gender. Both BAR and CelebA are heavily biased and contain no or very few conflicting examples--eg. CelebA Hair has only 1\% of men with blond hair in the train set. 

For BAR, since no validation set is provided, we show results both using in-distribution and out-distribution validation sets to compare against both sets of baselines: those that do and do not require conflicting examples in the validation set\footnote{For instance, methods that work on the principle of reweighting conflicting samples in the trainset (eg LWBC \cite{kim2022learning}) typically add 1\% of conflicting samples from the test set to the training data}. We outperform baselines in both settings by more than 1-2\% absolute accuracy, refer Table \ref{bar-table}.
Table \ref{celeba-table} shows results on CelebA dataset, we get almost the same unbiased accuracy as LWBC and improve upon conflicting accuracy.

\textbf{Domain-shift Generalization:} NICO introduces three new contexts in which object classes appear in the validation and test set, that are absent in the training set. Table \ref{nico-table} shows that \ouralgo\ beats all baselines on classification accuracy for the test set; this is despite our not using context information, unlike a majority of the baselines. Thus, \ouralgo\ is valuable for zero-shot domain generalization.

\textbf{Robustness to Texture bias}: \cref{imagenet9-table} shows results on ImageNet-9, which is known to be biased towards texture, and ImageNet-A, which consists of natural images that have bias-conflicting features. This setting is closest to real-world scenarios for texture bias. We improve the previous best baseline by 3\% absolute on ImageNet-9 validation set and by 4\% absolute on ImageNet-A test set. Thus, \ouralgo\ encourages learning features robust to texture bias, improving performance on both the in-distribution validation set as well as bias-conflicting test set. Two critical findings here are a) that \ouralgo\  did not sacrifice in-distribution accuracy through the process of sieving simple features, and b) the learned classifier robustly transfers over to a novel test set, where it provides even larger gains.

\begin{figure*}[!ht]
\vskip 0.2in
\begin{center}
\centerline{
    \includegraphics[width=.9\linewidth]{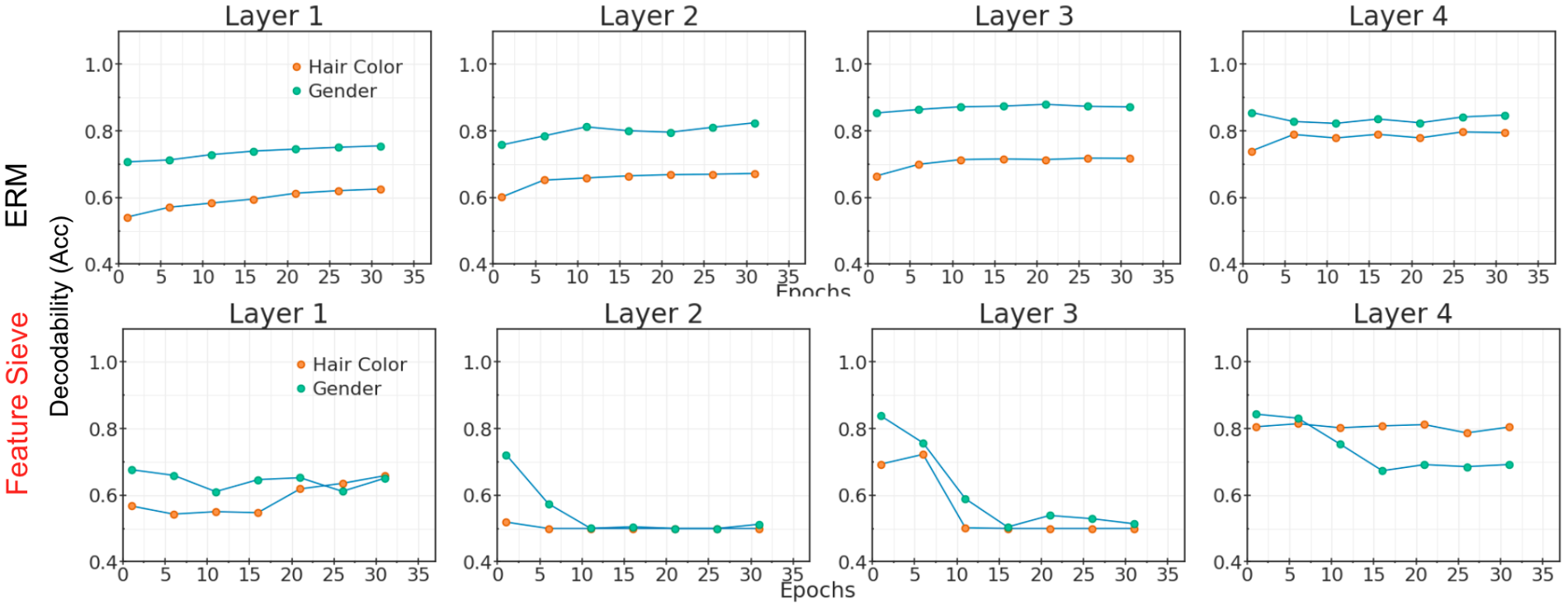}
}
\caption{Decodability of Spurious (Gender) and Target (Hair Color) features across layers of ResNet-18 while training on CelebA with a) Normal ERM training and b) with SIFER.
}
\label{fig:celeba_decodability}
\end{center}
\vskip -0.2in
\end{figure*}

\subsection{\ouralgo: Debiasing without Extra Information} 

\cref{bar-table} (BAR dataset, bottom half) shows that \ouralgo\ with only in-distribution validation data (cf. \cref{dataset-composition-table}) outperforms most baselines that leverage an additional OOD validation set (top half). Further, without using either attribute knowledge or conflicting examples in the validation set, we show huge gains over ERM (65.7\% accuracy vs 35.3\%), demonstrating that \ouralgo\ \textit{does not critically depend on} such additional information for debiasing, although we can certainly leverage such information for additional gains (72.8\% vs 65.7\% accuracy when using OOD validation data).

\subsection{Feature Decodability in Real-world Datasets}
Figure \ref{fig:celeba_decodability} shows feature decodability (\cref{sec:decode}) on the real world dataset CelebA, where the target label is hair color, and previous work has shown that gender is a spuriously correlated attribute. Results show that SIFER suppresses gender decodability, particularly in upper layers, with color feature achieving stronger decodability (unlike ERM where gender is more easily decodable than color). This mirrors the results on synthetic datasets that were presented in \cref{fig:cifar_mnist_decodability}, and shows that \ouralgo\ can \textit{automatically identify and suppress} featural information related to abstract concepts such as gender, in support of better generalization accuracy.

\subsection{\ouralgo\ focuses on Relevant Information}
We visualize the information in an image that is relevant to a given classifier~\cite{selvaraju2017grad}, in order to verify whether our feature sieving results in semantically relevant modifications to learned classifiers. \cref{fig:gradcam} shows this evaluation, contrasting ERM classifier's regions of focus (middle row) and \ouralgo's regions of focus (bottom row) on a range of input images (top row, drawn from BAR \& NICO). Interestingly, not only does \ouralgo\ correctly focus on the central object of interest, but also it is able to effectively suppress the (spuriously label-correlated) background information, which is highly valued by the ERM classifier. This undercores \ouralgo's ability to carefully differentiate between relevant and irrelevant features, rather than some notion of simple vs complex features alone.

\section{Discussion \& Conclusion}

We proposed \ouralgo--a novel \textit{feature sieve} approach towards addressing simplicity bias and spurious correlations in deep neural networks. Our proposal introduces an auxiliary network attached to the deep network which alternately identifies and suppresses predictive features. The approach is controllable through the use of configuration parameters optimized using validation data; thus, it requires no foreknowledge or hand-coding of the notion of ``simple features''. We demonstrated on controlled datasets the ability of \ouralgo\ to automatically identify and suppress features; further, we showed that, strictly speaking, \ouralgo\ \textit{rebalances} the role of various features in a controllable manner driven by the needs of generalization. We showed using extensive experiments on real-world data that our approach provides significant gains--3-11\% relative accuracy improvements on BAR, NICO, and Imagenet-A. We believe our work is a small, important first step in a fruitful new direction
of research.  We hope that follow-up work will build on the notion of the feature sieve, developing effective computational barriers that encourage deep networks to discover and utilize richer, more powerful featural representations. Our current approach strikes a balance between  various competing features, guided by generalization error estimates (validation error). One could potentially extract even more value if different feature classes could be isolated into (relatively) independent predictors, then combined effectively. This is, for instance, the approach taken by~\citet{niu2022roadblocks}. Thus, a straightforward next step we aim to explore is the study of ensembling approaches to combine a range of features of varying complexity \& predictive power, and methods for efficiently learning them.  We also hope to develop a systematic theoretical understanding of feature sieve approaches and their role in supervised learning using DNNs.




\bibliography{simbias}
\bibliographystyle{icml2022}


\newpage
\appendix
\onecolumn
\leftline{ {\Large Appendix } }

\section{Additional Details}
\subsection{Training Details}
\label{app-training_details}
For all experiments we consistently used ResNet-18, an auxiliary layer that uses the same layer structure as of BasicBlock of ResNet with varying depth. The ResNet network is composed of 4 layers modules (each itself is made up of 2 BasicBlocks). We apply auxiliary layer only at the end of the of layers except layer 4, this gives us 3 different choice for aux position($\mathcal{A}_P$) which we treat as hyperparameter. The network is optimized using SGD optimizer with a fixed learning rate of 0.001.  For real-world experiments the model is loaded with ImageNet pre-trained weights. We repeat the experiments with 5 different random seeds and report the mean and std deviation of results.
Table \ref{hparam-table} shows the hyperparamters search space for all the hyperparameters that we tune on the basis of validation set. To reduce hyperparameter search space we fixed the value of $\alpha1$ to 10. Table \ref{best-hparam-table} shows the hyperparameter values obtained from the hyperparamter tuning.

\begin{table}[h]
\caption{Range for hyperparameters search.}
\label{hparam-table}
\begin{center}
\begin{tabular}{lcc}
\toprule
\multicolumn{1}{c}{\textbf{Hparam}} & \textbf{Range} &  \\ \midrule
$\mathcal{A}_D$                           & $[1, 9]$                             &  \\
$\mathcal{A}_P$                             & $[1, 3]$                             &  \\
$\alpha2$                              & loguniform$(10^{-1}, 10^2)$                 &  \\
$\alpha3$                              & loguniform$(10^{-1}, 10^2)$                 &  \\
$\mathcal{F}$                             & $[1, 9]*10$                          & \\ \bottomrule
\end{tabular}
\end{center}
\end{table}

\begin{table}[h]
\caption{Hyperparameter values obtained from the tuning.}
\label{best-hparam-table}
\begin{center}
\begin{tabular}{lccccc}
\toprule
\textbf{Dataset}      & \textbf{$\mathcal{A}_D$} & \textbf{$\mathcal{A}_P$} & \textbf{$\alpha2$} & \textbf{$\alpha3$} & \textbf{$\mathcal{F}$}  \\ \midrule
BAR - ID val & 4  & 2  & 2      & 4.5    & 70 \\
BAR - OD val & 2  & 2  & 1      & 3      & 30 \\
CelebA       & 2  & 2  & 25     & 15     & 50 \\
NICO         & 2  & 1  & 1      & 75     & 70 \\
IN-9/IN-A    & 4  & 3  & 1      & 4.5    & 70 \\ \bottomrule
\end{tabular}
\end{center}
\end{table}

\section{Baselines}
Here we list and briefly explain all the baselines that we compare against on real world datasets: \\
\textbf{BiaSwap}~\cite{kim2021biaswap} proposes a bias-tailored augmentation-based approach for learning debiased representation without requiring supervision on the bias type. they divide the data into bias-guiding and bias-conflicting groups and then swaps the bias in bias guiding group.\\
\textbf{LfF}~\cite{nam2020learning} uses generalized cross-entropy initially trains a prejudiced net-work and tries to debias the second network by focusing weighing on samples that go against the bias. \\
\textbf{IRM}~\cite{arjovsky2019invariant} uses theory of causal bayesian networks to find an invariant feature representation using multiple training environments with different bias correlations.\\
\textbf{REx}~\cite{krueger2021out} proposed a min-max algorithm to optimize for the worst linear combination of risks on different environments.\\
\textbf{EIIL}~\cite{creager2021environment} optimizes for bias group assignment to automatically identify the bias groups to maximize IRM.\\
\textbf{PGI}~\cite{ahmed2021systematic} follows EIIL to identify bias groups by training a small neural network.\\
\textbf{Evading Simplicity Bias (ESB)}~\cite{teney2022evading} creates a ensemble of diverse classifiers by incorporating a diversity regularizer between the gradients while training.\\
\textbf{Roadblock}~\cite{niu2022roadblocks} adds adversarial augmentations to the image while training to avoid over-reliance on spurious visual cues.\\
\textbf{Debian}~\cite{li2022discover} trains two networks in alternate manner namely discoverer and classifier, the discoverer tries to find multiple unknown biases of the classifier without any annotations of biases, and the classifier aims at unlearning the biases identified by the discoverer.\\
\textbf{ReBias}~\cite{bahng2020learning} propose a novel framework to train a de-biased representation by encouraging it to be different from a set of representations that are biased by design.\\
\textbf{LWBC}~\cite{kim2022learning} employs a committee of classifiers as an auxiliary module that identifies bias-conflicting data and assigns large weights to them when training the main classifier.\\
\textbf{Group-DRO}~\cite{sagawa2019distributionally} minimizes for worst-case training loss over a set of pre-defined groups.\\
\textbf{EnD}~\cite{tartaglione2021end} proposes a regularization technique that uses the bias attributes to prevent deep models from learning spurious biases by inserting an information bottleneck. \\
\textbf{CSAD}~\cite{zhu2021learning}, given the bias attributes, explicitly extracts target and bias features disentangled from the latent representation generated by a feature extractor and then learns to discover and remove the correlation between the target and bias features.\\
\textbf{JiGen}~\cite{carlucci2019domain} jointly classifies objects and solves unsupervised jigsaw tasks.\\
\textbf{Cumix}~\cite{mancini2020towards} mixes up data and labels from different domains to be able to recognize unseen categories in unseen domains.\\
\textbf{MTL}~\cite{blanchard2021domain} argue that problem of Domain Generalization can be viewed as a kind of supervised learning problem by augmenting the original feature space with the marginal distribution of feature vectors.\\
\textbf{DANN}~\cite{ganin2016domain} proposes a representation learning approach such that features are not predictive of the domain from which the model is being trained on.\\
\textbf{CORAL}~\cite{sun2016deep} proposes an unsupervised domain adaptation method that aligns the second-order statistics of the source and target distributions with a linear transformation.\\
\textbf{MMD}~\cite{li2018domain} extend adversarial autoencoders by imposing the Maximum Mean Discrepancy measure to align the distributions among different domains, and matching the aligned distribution to an arbitrary prior distribution via adversarial feature learning.\\
\textbf{CNBB}~\cite{he2021towards} is an OoD learning method that based on sample reweighting inspired by causal inference.\\
\textbf{DecAug}~\cite{bai2020decaug} proposed a semantic augmentation and feature decomposition approach to distangle context features from category related features.\\
\textbf{NAS-OoD}~\cite{bai2021ood} adds an OOD generalization criterion to network architecture search training to construct inherently more robust network architectures.\\
\textbf{StylisedIN}~\cite{geirhos2018imagenet} showed that ImageNet is texture biased and works on improving shape bias.\\
\textbf{LearnedMixin}~\cite{clark2019don} trains a robust model as part of an ensemble with the naive one in order to encourage it to focus on other patterns in the data that are more likely to generalize.\\
\textbf{CaaM}~\cite{wang2021causal} learns causal attention by partitioning the data on-the-go to break correlation with bias.\\


\end{document}